\documentclass[letterpaper]{article} % DO NOT CHANGE THIS
\usepackage{aaai2026}  % DO NOT CHANGE THIS
\usepackage{times}  % DO NOT CHANGE THIS
\usepackage{helvet}  % DO NOT CHANGE THIS
\usepackage{courier}  % DO NOT CHANGE THIS
\usepackage[hyphens]{url}  % DO NOT CHANGE THIS
\usepackage{graphicx} % DO NOT CHANGE THIS
\usepackage{natbib}  % DO NOT CHANGE THIS AND DO NOT ADD ANY OPTIONS TO IT
\usepackage{caption} % DO NOT CHANGE THIS AND DO NOT ADD ANY OPTIONS TO IT
\usepackage{algorithm}
\usepackage{algorithmic}
\usepackage{booktabs}
\usepackage{multirow}
\usepackage{graphicx}

\usepackage{array}

\usepackage{newfloat}
\usepackage{listings}
\DeclareCaptionStyle{ruled}{labelfont=normalfont,labelsep=colon,strut=off} % DO NOT CHANGE THIS
\floatstyle{ruled}
\newfloat{listing}{tb}{lst}{}
\floatname{listing}{Listing}
\title{CL-DMDF: Dynamic Multimodal Data Fusion Model \\Based on Contrastive Learning}
\author{
    Dong Li\textsuperscript{\rm 1},
    Lingling Zhang\textsuperscript{\rm 1},
   Binghao Han\textsuperscript{\rm 1},
   Linlin Ding\textsuperscript{\rm 1}\thanks{Corresponding author},
    Yue Kou\textsuperscript{\rm 2}
}
\affiliations{
    \textsuperscript{\rm 1}School of Information, Liaoning University, Shenyang, China\\
    \textsuperscript{\rm 2}School of Computer Science and Engineering, Northeastern University, Shenyang, China
   dongli@lnu.edu.cn, \{4032532425, 4032232396\}@smail.lnu.edu.cn, dinglinlin@lnu.edu.cn, kouyue@cse.neu.edu.cn
}

\usepackage{bibentry}
\begin{document}

\maketitle

\begin{abstract}
Multimodal data fusion involves integrating and analyzing information from multiple modalities to uncover latent correlations and complementary patterns, thereby enhancing data processing and decision-making. While existing methods for structured multimodal inputs are typically designed around specific tasks and assume fully observed modalities, real-world applications often suffer from uncertain or missing modality inputs due to various factors. Some traditional models overly emphasize local interactions within missing modalities, neglecting the global complementary cues embedded in multimodal representations. To overcome these limitations, we propose a Dynamic Multimodal Data Fusion model based on Contrastive Learning (CL-DMDF). CL-DMDF introduces a novel attention mechanism that operates across both feature and modality dimensions to compute reliable attention scores, effectively reflecting importance at each level. The CL-DMDF further incorporates an entity-centroid contrastive learning module that constructs centroid-based positive samples from entity features to enhance discriminative learning. Additionally, an adaptive fusion module is employed to improve the efficiency and accuracy of dynamic fusion strategies. Extensive experiments conducted on three datasets demonstrate the effectiveness of the CL-DMDF across diverse multimodal fusion tasks.
\end{abstract}

\begin{links}
    \link{Code}{https://github.com/zoo-111-p/CL-DMDF}
    % \link{Datasets}{https://aaai.org/example/datasets}
    % \link{Extended version}{https://aaai.org/example/extended-version}
\end{links}

\section{Introduction}

In the real world, humans perceive the environment through sensory receptors such as eyes, ears, skin, nose, and tongue, enabling them to see objects, hear sounds, feel textures, smell scents, and taste flavors. The information obtained from each sensory source or medium can be regarded as a modality. Multimodal refers to the use of two or more heterogeneous modalities, such as text, vision, or audio, for joint learning and inference.

The human brain unconsciously integrates information from different sensory receptors, or fuses modalities, extracting complementary information to form predictions or decisions. In addition, machines are highly dependent on sensors such as RGB cameras, microphones, and other types of sensors. Each sensor maps the observed objects or activities into the machine's domain, enabling it to make predictions or decisions based on the collected data.

While existing methods for multimodal decision-making often rely on manual feature engineering, they often fail to capture cross-modal complementarity, leading to early information loss. Although models such as CNNs, LSTMs, transformers, and BERT can process multimodal inputs, the key challenge remains how to perform deep fusion to enhance machine intelligence.

\begin{figure}[t]
\centering
\includegraphics[width=1\columnwidth]{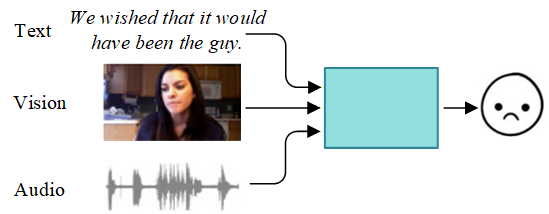} 
\caption{Relying solely on textual information is unlikely to accurately predict the current emotion during classification. However, audio and vision modalities can provide crucial cues for the multimodal network.\cite{xue2023dynamic}}
\label{fig1}
\end{figure}

Multimodal data fusion aims to combine information from multiple modalities—such as vision, audio, and textual—to enhance analytical accuracy. For example, in emotion recognition (Figure 1), integrating audio and visual signals improves classification by leveraging cross-modal complementarity.

For dynamic multimodal data, we propose a Dynamic Multimodal Data Fusion model based on Contrastive Learning (CL-DMDF). The model extracts modality-specific features and projects them into a shared embedding space. A dual-dimensional attention mechanism over feature and modality dimensions assigns weights to account for structural variability. Contrastive learning enhances feature discriminability, while an adaptive fusion module dynamically selects task-relevant strategies to generate a unified representation.

Multimodal fusion significantly enhances system perception by mitigating the limitations of single-modality inputs, which are often affected by factors such as lighting, noise, and sensor failure. By integrating complementary cues, it improves accuracy and robustness in complex environments. For instance, in autonomous driving, combining camera images, radar data, and voice commands enhances safety and interaction reliability.

In this paper, we propose a novel Dynamic Multimodal Data Fusion model based on Contrastive Learning (CL-DMDF), to address the limitations of existing fusion models in terms of task adaptability and semantic expressiveness. Our main contributions are summarized as follows:

(1) We propose a dual-dimensional attention mechanism that jointly models feature-level and modality-level importance, enabling the computation of more reliable attention scores to guide effective multimodal integration.

(2) We introduce an entity-centroid contrastive learning module, which constructs positive and negative pairs based on attention-weighted modality features. This module enhances the discriminative power of representations and expands the embedding space.

(3) We present an adaptive fusion module that dynamically selects optimal fusion strategies based on the characteristics of input features. This allows the model to balance accuracy and computational efficiency across varying tasks.

(4) CL-DMDF's innovation lies in the collaborative design of its three components, optimizing a single objective within a unified architecture, representing an innovation at the framework level. We conduct comprehensive experiments on three representative datasets. The results demonstrate that CL-DMDF consistently outperforms strong baselines across diverse tasks, validating the effectiveness and generalizability of the model.

\section{Related Work}
Multimodal data fusion under dynamic conditions aligns with conventional fusion paradigms. This section reviews representative studies in this area.

\subsection{Multimodal Data Fusion}
Multimodal fusion has traditionally been categorized into data-level, feature-level, and decision-level approaches. Early data-level models \cite{camille2013} concatenated RGB and depth images at the input, while SSR-CNN \cite{liu2019} employed a single-stream architecture to integrate modalities. Feature-level approaches \cite{li2018, hu2020} encoded modalities independently and fused them during decoding; FFN \cite{janani2021} improved this by concatenating outputs from modality-specific encoders.

For decision-level fusion, \cite{nihar2021} introduced a multimodal variational autoencoder (VAE) that learned a shared latent space from image features. Similarly, an end-to-end VAE framework \cite{dhruv2019a} addressed fake news detection by encoding and reconstructing joint textual and visual embeddings.

Recent studies extended these paradigms. \cite{chen2024pcam} proposed a progressive cross-modal attention mechanism for adaptive fusion across abstraction levels. \cite{liu2025lvmf} incorporated latent-variable modeling to capture modality-specific uncertainty. \cite{zhang2025lightfusion} further explored lightweight transformer-based fusion under real-time constraints, improving the trade-off between efficiency and performance.

\subsection{Dynamic Multimodal Data Fusion}
Dynamic attention-based fusion methods were generally categorized into intra-modality self-attention, cross-modality cross-attention, and transformer-based approaches. \cite{gao2019} applied hard attention to generate spatial binary masks for selective feature propagation. \cite{mateusz2018} introduced bidirectional cross-modal attention for vision-language alignment, and \cite{hu2020b} proposed a dot-product cross-attention to capture audio-text correlations. MVAE integrated multiple modalities for tasks such as fake news detection.Transformer-based fusion methods leveraged cross-modal attention to model long-range dependencies. \cite{sun2021} designed a cross-modal transformer for MRI–acoustic signal alignment, while \cite{xu2022} adopted self-attention to capture inter-modal relations. \cite{yang2024dynamic} presented a multi-head sparse transformer with hierarchical fusion, demonstrating robustness on noisy video-text datasets. \cite{liu2025exploring} introduced an adaptive sparse attention framework that pruned modality contributions based on semantic uncertainty. More recently, UniFM~\cite{jiang2024unifm} and MM-TokenMixer~\cite{li2025tokenmixer} optimized cross-modal token integration using shared representations and token-wise mixing, resulting in improved generalization between benchmarks.Dynamic multimodal fusion has also been investigated to improve the efficiency of multimodal inference~\cite{xue2023dynamic}, which motivates further exploration of flexible fusion strategies under complex multimodal settings.

Graph-based methods evolved from standard GCNs to spatio-temporal graph architectures. \cite{chih2022} utilized deep GCNs for emotion recognition, and \cite{hu2021} combined multimodal GATs with temporal convolutions to model temporal-spatial patterns. \cite{ding2023} integrated multi-head attention into GNNs for scene graph embeddings, followed by cross-modal alignment in \cite{yang2023}. \cite{li2023incorporating} built a movie KG from metadata and used its embeddings in multimodal genre classification via self-supervised attention and contrastive learning, showing structured relational knowledge complements multimodal representation learning. \cite{wang2025modality} proposed a reinforcement-guided GNN with modality-aware policy learning for dynamic social event detection. Furthermore, GNN-Adapter~\cite{wu2024gnnadapter} incorporated lightweight graph modules into pre-trained multimodal models, enhancing efficiency without retraining the backbones.

% Graph-based methods evolved from standard GCNs to spatio-temporal graph architectures. \cite{chih2022} utilized deep GCNs for emotion recognition, and \cite{hu2021} combined multimodal GATs with temporal convolutions to model 
% temporal-spatial patterns. \cite{ding2023} integrated multi-head attention into GNNs for scene graph embeddings, followed by cross-modal alignment in \cite{yang2023}. Recent work has also explored the use of domain knowledge graphs for multimodal classification. \cite{li2023incorporating} constructed a movie-domain knowledge graph from metadata and incorporated knowledge graph embeddings into multimodal movie genre classification with self-supervised attention and contrastive learning.This suggests that structured relational knowledge can provide complementary cues for multimodal representation learning. \cite{wang2025modality} proposed a reinforcement-guided GNN with modality-aware policy learning for dynamic social event detection. Furthermore, GNN-Adapter \cite{wu2024gnnadapter} incorporated lightweight graph modules into pre-trained multimodal models, enhancing efficiency without retraining the backbones.

\subsection{Differences from Existing Work}

\begin{figure*}[t]
\centering
\includegraphics[width=\textwidth]{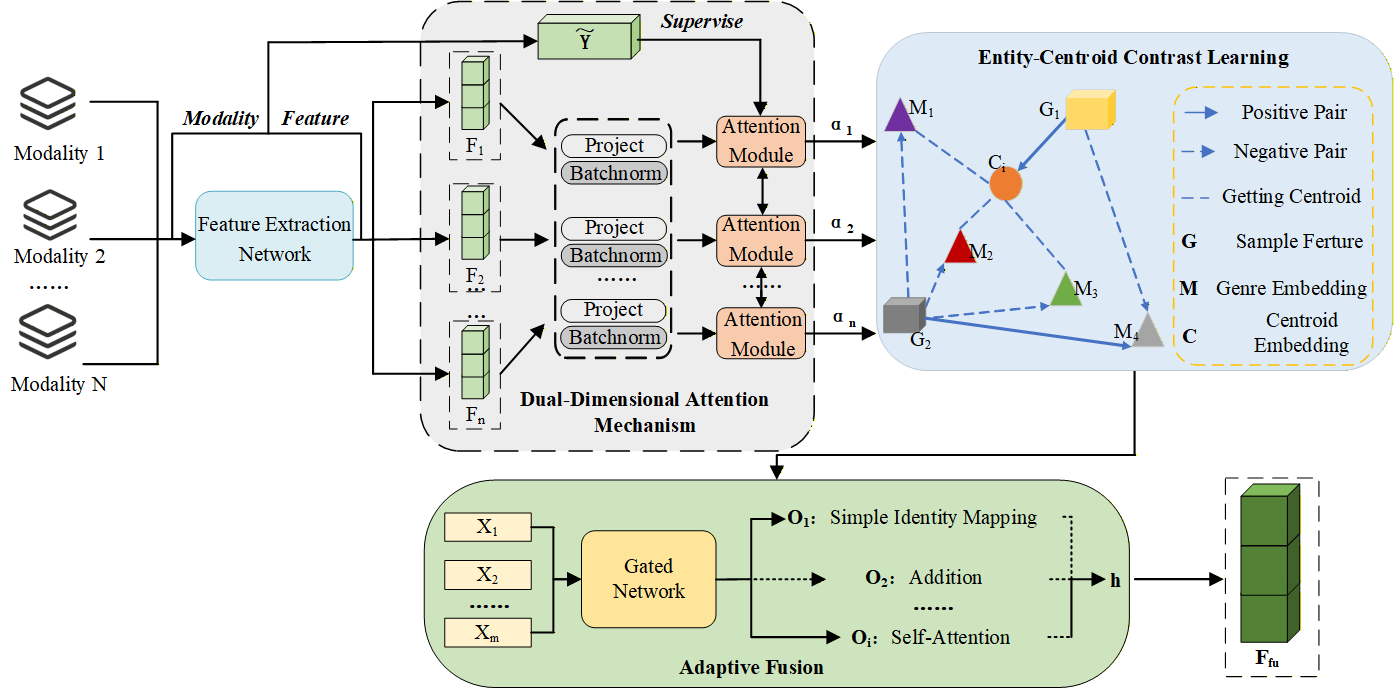}
\caption{The overview of CL-DMDF. Features are first extracted from data of different modalities using a feature extraction network and projected into a unified dimensional space. A dual-dimensional attention mechanism is then employed to guide the allocation of attention. Subsequently, contrastive learning is applied to enhance the discriminability of the features. Finally, an adaptive fusion module selects the most appropriate fusion strategy based on the specific requirements of the task.}
\label{fig2}
\end{figure*}

Our work differs from traditional methods in several key aspects.

Most existing fusion models adopt static strategies across tasks, limiting adaptability. In contrast, CL-DMDF employs a dynamic fusion mechanism that selects task-relevant strategies based on modality reliability. Existing attention-based methods often yield unstable weights due to unsupervised design, while graph-based models struggle to generalize to unseen modalities. CL-DMDF addresses this via a dual-dimensional attention mechanism that jointly considers feature-level and modality-level importance. Unlike previous contrastive learning approaches with coarse alignment, our entity-centric contrastive module captures fine-grained semantics, enhancing representation quality. Finally, to balance performance and efficiency, we introduce a resource-aware objective that guides the adaptive fusion module to avoid redundant computation under varying task complexities.

\section{Method}
We begin by presenting the proposed CL-DMDF model for dynamic multimodal fusion, followed by a formal definition of the problem and its implementation details.

\subsection{Model Overview}
This paper proposes a Dynamic Multimodal Data Fusion model based on Contrastive Learning, termed CL-DMDF. The model first extracts features from different modalities using dedicated feature extraction networks and projects them into a unified vector space. To account for the diversity of modality combinations and varying feature counts per entity, a dual-dimensional attention mechanism is introduced to guide attention allocation across modalities and features. This mechanism enhances the model's ability to concentrate on task-relevant information. To enhance feature discriminability, CL-DMDF integrates a contrastive learning module that sharpens representation boundaries by distinguishing similar from dissimilar samples. An adaptive fusion module further selects optimal strategies based on task demands and modality characteristics, enabling effective aggregation of heterogeneous features. The overall architecture is shown in Figure 2.

\subsection{Dual-Dimensional Attention Mechanism} 
Different modalities may represent distinct entities with varying semantic features, complicating consistent and informative fusion due to modality-specific encoding differences.To address this, we propose a dual-dimensional attention mechanism that jointly considers both feature-level richness and modality-level presence when assigning attention weights to each entity. 

Entities with broader feature coverage and presence across more modalities are assigned higher weights, enhancing their contribution to the final fused representation. This method prioritizes semantically important entities, enabling more effective cross-modal integration. Weighted features are used in contrastive learning, with an attention module, consisting of a shared linear layer and non-linear activation, further balancing modality contributions across samples.

For multimodal features, \(F^1_i \in R^{D_1}, F^2_i \in R^{D_2}, \dots F^n_i \in R^{D_n}\), where \( \{D_i\}\) represents the features extracted from different modalities, we use batch normalization and linear projection functions to convert them into the same shape, is given by Equation 1:

\begin{equation}\label{eq:proj}
  h(x)=\mathrm{project}\big(\mathrm{batchnorm}(x)\big)
\end{equation}

where \( project(\cdot) \) and \( batchnorm(\cdot) \) represent the linear projection function and batch normalization function, respectively.
After alignment, all modality features are input into the module to obtain their attention scores \( a_i \), which guide the contrastive learning process to generate centroids and expand the feature embedding space.

To improve attention reliability, we introduce a self-supervised dual-dimensional mechanism that assigns attention scores based on pseudo-labels \(\tilde{Y}_n\), reflecting both feature count and modality coverage. Entities with limited features or modalities receive lower weights due to reduced semantic value. The pseudo-label \(\tilde{Y}_n\) is defined over $F$ as shown in Equations 2-4.

\begin{equation}\label{eq:N}
  \overline{N} = \frac{1}{W}\sum_{i=1}^{W} F^{i}
\end{equation}

\begin{equation}\label{eq:M}
  \overline{M} = \frac{1}{W}\sum_{i=1}^{W} \sum_{E_i \in M_j} M_{E^i}
\end{equation}

\begin{equation}\label{eq:Y}
  \tilde{Y} = \frac{N_i \sum_{E^i \in M_j} M_{E^i}}{N_i + \overline{N} \left( \sum_{E^i \in M_j} M_{E^i} + \overline{M} \right)}
\end{equation}

Let \(W\) denote the number of entities and \(M_{E^i}\) the number of modalities containing entity \(E^i\). Entities with feature and modality counts above the averages \(\overline{N}\) and \(\overline{M}\) are assigned higher pseudo-labels \(\tilde{Y}_n\), while others receive lower values. To align with the activation-constrained attention range, \(\tilde{Y} \in \{0,1\}\) is used.

To ensure stable convergence of self-supervised attention scores, we incorporate an appropriate loss function as shown in Equation 5.

\begin{equation}\label{eq:Gamma}
  \Gamma_{\mathrm{atten}} = - \sum_{i=1}^{B} \log \left( 1 - |\alpha_{M_i} - \overline{Y_i}| \right)
\end{equation}

Since \( \alpha_{M_i} \) is a vector and \( \tilde{Y}_n \) is a scalar, we first average \( \alpha_{M_i} \) to obtain a scalar for regression. As both values lie within \( (0,1) \), directly applying standard \( L_1 \) or \( L_2 \) losses can lead to weak gradients. To address this, we adopt a logarithmic transformation in the loss function to maintain sufficiently strong gradients and ensure monotonicity during optimization.

We align features using batch normalization and linear projection to stabilize attention weights. CL-DMDF demonstrates robustness under noisy and low-sample conditions by focusing on key entities and enhancing features via two-dimensional attention and contrastive learning.

\subsection{Entity-Centroid Contrastive Learning}

\begin{figure}[t]
\centering
\includegraphics[width=1\columnwidth]{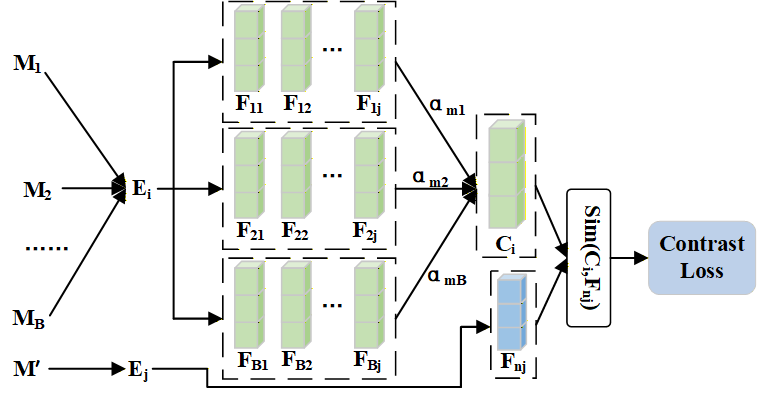}
\caption{Entity-Centroid Contrastive Learning implementation details. Embedding features in vector space to capture diverse semantics and prevent over-alignment in contrastive learning.}
\label{fig3}
\end{figure}

Figure 3 illustrates the Entity-Centroid Contrastive Learning framework. To enhance discriminability, representations are projected into a shared embedding space. Unlike conventional methods, our approach captures multiple semantics per instance. For a batch of fused features \(F^{Mi}\), each entity aggregates cross-modal information, with its embedding set defined in Equation 6.

\begin{equation}\label{eq:G}
  G_i =  F^{M_1}, F^{M_2}, \dots, F^{M_n} 
\end{equation}

Let \(n\) denote the number of modalities in the fused feature set. The centroid \(C_i\) of group \(G_i\) expands the embedding space. Dual-dimensional attention scores guide weighted feature aggregation, integrating modality-specific information, with features treated as positive samples (Equation 7).

\begin{equation}\label{eq:C}
  C_i = \sum_{F^{M_i} \in G_i} \alpha_{M_i} F^{M_i}
\end{equation}

Here, \(\alpha_{M_i}\) denotes the attention scores generated by the dual-dimensional attention mechanism. After expanding the embedding space, features within this space are treated as positive samples, while those outside it are considered negative. Accordingly, the complete embedding set for all samples is defined as in Equation 8.

\begin{equation}\label{eq:U}
  U^B_{i=1}{G_i} = \{ F^{M_1}, F^{M_2}, \dots, F^{M_j} \}
\end{equation}

\( B \) denotes the number of entities, and \( j \) the total number of modality features. This yields the feature embedding set for all entities in the current batch. Samples outside this embedding space—i.e., the complement of \( U_{i=1}^BG_{i} \) is considered a negative sample during reinforcement learning, as in Equation 9.

\begin{equation}\label{eq:S}
  S_i = U^B_{i=1}{G_i} - C_i
\end{equation}

In the loss design, since positive samples define an embedding space, we compute the similarity between each negative sample and all features within this space and sum the results to enhance feature discriminability, which can be formulated as follows:

\begin{equation}\label{eq:q}
  q_i = \sum_{F^{M_i} \in S_i} \exp\left( \frac{F_{i}f_{E^{M_i}}}{\tau} \right)
\end{equation}

\begin{equation}\label{eq:Gamma_contra}
  \Gamma_{\mathrm{contra}} = - \sum_{i=1}^{B} \log \frac{\exp\left( \frac{F_{i} f_{C_i}}{\tau} \right)}{{\exp\left( \frac{F_{i} f_{C_i}}{\tau} \right)} + q_i}
\end{equation}

The temperature coefficient \(\tau\) adjusts the model's sensitivity to negative samples. Lower \(\tau\) sharpens similarity scores, assigning higher penalties to hard negatives. During training, negatives are repelled proportionally to their similarity. \(q_i\) denotes the total similarity between sample \(F_i\) and its positives \(S_i\), with the objective of attracting positives and repelling negatives.

Given multimodal inputs \(X = \{x_1, x_2, ..., x_M\}\), we define a set of candidate fusion operations \(\{O_i\}_{i=1}^{B}\), such as element-wise addition, concatenation, or attention-based fusion. A gating network \(G(\cdot)\) processes \(X\) and outputs a weight vector of \(B\). The final fused representation \(h\) is computed as the weighted sum over all fusion operations, as in Equation 12.

\begin{equation}\label{eq:h}
  h = \sum_{i=1}^{B} g_i o_i(x)
\end{equation}

The Entity-Centric Contrastive Learning module tackles class imbalance by using attention-weighted features for positives and out-of-embedding samples for negatives.

 \subsection{Adaptive Fusion}

The adaptive fusion module integrates multiple fusion operations with a gating mechanism that dynamically selects task-relevant strategies based on contrastively enhanced, weighted features, enabling flexible multimodal fusion.

\begin{figure}[t]
\centering
\includegraphics[width=1\columnwidth]{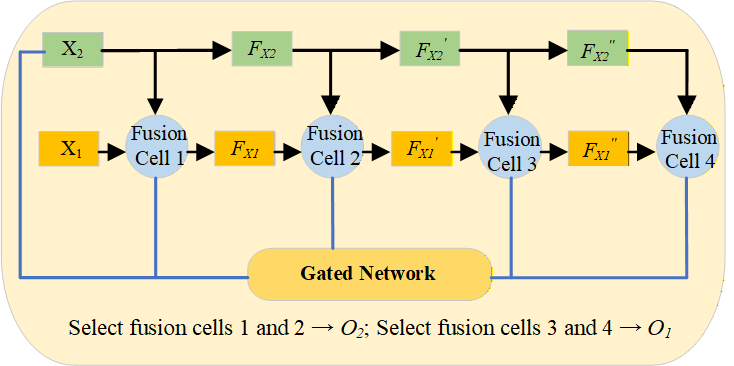} 
\caption{The adaptive fusion module enables finer-grained and more flexible decision-making by stacking multiple fusion units to construct a fusion network.}
\label{fig4}
\end{figure}

As shown in Figure 4, $X_1$ and $X_2$ are inputs from two modalities. The module includes four fusion blocks and a global gating network, where $F_r$, $F_t$ are shallow features, and $F_r'$, $F_t'$, $F_r''$, $F_t''$ are deeper features. The gating network selects fusion units based on task requirements.

To enhance efficiency, deeper operations are skipped when shallow features suffice. The gating network selects operations based on task complexity. A resource-aware loss is introduced, with $C(O_{i,j})$ representing the cost of operation $j$ in fusion unit $i$. During inference, the adaptive module dynamically adjusts fusion strategies based on input features, managing modality uncertainty through resource-aware optimization (Eq. 13).

\begin{equation}\label{eq:Gamma_fusion}
  \Gamma_{\mathrm{fusion}} = \Gamma_{\mathrm{task}} + \sum_{j=1}^{F} \sum_{i=1}^{B} g_i^{(j)} C(O_{i,j})
\end{equation}

Here, $\Gamma_{\mathrm{task}}$ denotes the task loss, $g_i^{(j)}$ is the policy weight assigned to the $j$-th operation by the $i$-th fusion cell, $B$ is the total number of candidate operations, and $F$ is the number of fusion cells. The final training objective of the CL-DMDF model combines the attention loss $\Gamma_{\mathrm{atten}}$ (Equation 5), contrastive loss $\Gamma_{\mathrm{contra}}$ (Equation 11), and fusion loss $\Gamma_{\mathrm{fusion}}$ (Equation 13), formally defined in Equation 14.

\begin{equation}\label{eq:Gamma}
  \Gamma = \Gamma_{\mathrm{atten}} + \Gamma_{\mathrm{contra}} + \Gamma_{\mathrm{fusion}}
\end{equation}

\subsection{Dynamic Multimodal Fusion Algorithm}%3.3.5
This section outlines the general CL-DMDF algorithm. Given a multimodal data set $D$, hyperparameter $\tau$, and iteration count $num$, the model initializes and outputs the fused feature representations. The computation is defined as follows.

\begin{itemize}
        \item \textbf{Step 1}: Initialize all modalities and their corresponding feature embeddings, then extract and embed modality-specific features from the dataset.
        \item \textbf{Step 2}: Extract modality features and enhance embeddings by minimizing a temperature-scaled contrastive loss between positive and negative pairs.
        \item \textbf{Step 3}: Compute attention scores based on feature and modality averages, and fuse features via the adaptive module to obtain the final representation.
\end{itemize}

For full implementation details, please refer to the \textbf{supplementary material}.

\section{Experiment}% (3.4实验结果与分析)
We evaluated CL-DMDF on three public datasets, with experiments including setup details, comparisons, and ablations.

\subsection{Experimental Setup}%3.4.1

\subsubsection{Datasets. }  
\textit{MM-IMDB} contains approximately 25{,}000 training and 25{,}000 testing movie reviews, each consisting of textual content and a corresponding movie poster image, labeled with either positive or negative sentiment. \textit{NYU Depth V2} includes 1{,}449 images captured from more than 400 indoor scenes, each with pixel-level depth annotations. \textit{CMU-MOSEI} consists of more than 23{,}000 video clips, each annotated with sentiment categories and containing data from multiple modalities.

To improve the accuracy of the model, we progressively adjust key hyperparameters during training. Selected experimental settings for CL-DMDF are summarized in Table 1. Please refer to the \textbf{supplementary material} for specific extractor configurations.

\subsubsection{Baseline Methods. }
To ensure fair evaluation across heterogeneous tasks, we adopt dataset-specific metrics. \textit{MM-IMDB}, a multi-label text-image classification dataset, uses \textit{Micro-F1} and \textit{Macro-F1}, capturing global and per-class performance, respectively. \textit{NYU Depth V2}, for semantic segmentation, is evaluated via \textit{Mean Intersection over Union (MIoU)}.  \textit{CMU-MOSEI}, a multimodal sentiment benchmark, reports \textit{Accuracy (Acc)} and \textit{Mean Absolute Error (MAE)} for classification and regression. 
This metric selection supports a comprehensive and task-aligned assessment of CL-DMDF.

We evaluate CL-DMDF on \textit{MM-IMDB}, \textit{NYU Depth V2}, and \textit{CMU-MOSEI}, comparing it against three fusion model categories and unimodal baselines. 

(1)\textbf{Unimodal Models}: Text-only and image-only baselines are included to assess the standalone discriminative power of individual modalities.

(2) \textbf{Encoder-Decoder Models}: CCA~\cite{sun2020}, MFM~\cite{braz2021}, ReFNet~\cite{sankaran2021}, MUIT~\cite{bai2019}, CM-BERT~\cite{yang2020}, and CEN~\cite{wang2020}; 

(3) \textbf{Attention-Based Models}: MulCon~\cite{chih2022}, MARN~\cite{amir2018}, LW-RefineNet~\cite{vladimir2018}, and ESANet~\cite{daniel2021}; 

(4) \textbf{Graph-Based Models}: MLTC~\cite{wang2022} and MI-Matrix~\cite{siddhant2019}, which take advantage of graph reasoning for modality interaction.

The selected benchmarks cover diverse modalities (discrete, continuous, temporal), validating our model's ability to handle dynamic multimodal interactions and supporting future adaptation to unseen modalities.

\begin{table}[t]
\centering
\scriptsize
\begin{tabular}{p{2.0cm} p{1.6cm} p{1.2cm} p{1.6cm}} % 控制每列宽度
\toprule
\textbf{Parameter} & \textbf{MM-IMDB} & \textbf{NYU-V2} & \textbf{CMU-MOSEI} \\
\midrule
Epoch            & 500 & 500 & 500 \\
Batch size       & 32  & 32  & 32  \\
Learning rate    & 0.001 & 0.001 & 0.001 \\
Vector dimension & 200 & 200 & 200 \\
\bottomrule
\end{tabular}
\caption{Hyperparameter settings of CL-DMDF across different datasets.}
\label{table:parameters}
\end{table}

\begin{table}[t]
\centering
\footnotesize
\begin{tabular}{p{2.0cm} p{2.6cm} p{1.0cm} p{1.1cm}} % 控制每列宽度
\toprule
& \textbf{Model} & \textbf{MicroF1} & \textbf{MacroF1} \\
\midrule
\multirow{2}{*}{Unimodal}
& Unimodal Text  & 59.37 & 47.59 \\
& Unimodal Image & 40.31 & 25.76 \\
\midrule
\multirow{4}{*}{Encoder-Decoder}
& LRMF   & 58.95 & 50.73 \\
& CCA    & 60.31 & 50.45 \\
& MFM    & 56.44 & 48.53 \\
& ReFNet & 59.45 & 51.51 \\
\midrule
\multirow{2}{*}{Attention-Based}
& RMFE   & 58.67 & 49.82 \\
& DynMM  & 60.35 & 51.60 \\
\midrule
Graph-Based
& MI-Matrix & 55.87 & 46.77 \\
\midrule
& \textbf{CL-DMDF(ours)} & \textbf{63.25} & \textbf{53.28} \\
\bottomrule
\end{tabular}
\caption{Comparison of CL-DMDF with baseline methods on MM-IMDB dataset.}
\label{table2}
\end{table}

\subsection{Experiment Results}
\subsubsection{Comparative Experiments. }
To thoroughly evaluate the performance of CL-DMDF, we performed comparative experiments on the MM-IMDB, NYU Depth V2 and CMU-MOSEI datasets. The results are presented in Tables 2-4.

Table 2 summarizes the results of CL-DMDF on the MM-IMDB dataset. The unimodal text model outperforms the image-only model, highlighting the advantage of text features. Most baseline fusion methods lacking modality-aware adaptation offer limited improvement. DynMM achieves Micro-F1 and Macro-F1 scores of 60.35\% and 51.60\%, respectively, while CL-DMDF's adaptive fusion and contrastive learning methods improve by 2.9\% and 1.68\%, respectively, with Micro-F1 and Macro-F1 scores reaching 63.25\% and 53.28\%, respectively.

Table 3 summarizes CL-DMDF’s results on the NYU Depth V2 dataset, comparing it with state-of-the-art semantic segmentation methods. CL-DMDF achieves the highest MIoU (52.3\%), surpassing CEN by 1.2\% while using only 1/14 of its computational cost. While CEN’s dynamic fusion strategy incurs high resource overhead, CL-DMDF balances performance and efficiency through task-adaptive fusion, outperforming all baselines in both accuracy and resource use.

\begin{table}[t]
\centering
\footnotesize
\begin{tabular}{p{2.0cm} p{2.4cm} p{1.4cm} p{1.0cm}} % 控制每列宽度
\toprule
& \textbf{Model} & \textbf{MIoU (\%)} & \textbf{MAdds} \\
\midrule
\multirow{2}{*}{Encoder-Decoder} 
& ACNet & 48.3 & 126.2 \\
& CEN   & 51.1 & 618.3 \\
\midrule
\multirow{2}{*}{Attention-Based} 
& ESANet       & 50.6 & 56.9 \\
& LW-RefineNet & 43.6 & \textbf{38.5} \\
\midrule
Graph-Based
& MLTC & 50.4 & 147.6 \\
\midrule
& \textbf{CL-DMDF (ours)} & \textbf{52.3} & \textbf{43.4} \\
\bottomrule
\end{tabular}
\caption{Comparison of CL-DMDF with baseline methods on NYU Depth V2 dataset.}
\label{tab:segmentation}
\end{table}

\begin{table}[t]
\centering
\footnotesize
\begin{tabular}{p{2.1cm} p{2.4cm} p{1.2cm} p{1.1cm}} % 控制每列宽度
\toprule
\textbf{} & \textbf{Model} & \textbf{Acc (\%)} & \textbf{MAE} \\
\midrule
\multirow{4}{*}{Encoder-Decoder} 
 & MFM & 78.1 & 0.951 \\
 & CM-BERT & 84.5 & \textbf{0.729} \\
 & LRMF & 76.4 & 0.912 \\
 & MUIT & 83.0 & 0.871 \\
\midrule
Attention-Based & MARN & 77.1 & 0.968 \\
\midrule
Graph-Based & MLTC & 78.4 & 0.922 \\
\midrule
\textbf{} & \textbf{CL-DMDF (ours)} & \textbf{85.4} & \textbf{0.737} \\
\bottomrule
\end{tabular}
\caption{Comparison of CL-DMDF with baseline methods on the CMU-MOSEI dataset.}
\end{table}

Table 4 shows CL-DMDF’s results on the CMU-MOSEI dataset, comparing it with baselines, including CM-BERT. While CM-BERT improves performance through cross-modal fine-tuning, CL-DMDF outperforms with higher accuracy (85.4\%, +0.9\%) and competitive MAE (0.737) by leveraging adaptive feature weighting and contrastive centroid embeddings, demonstrating superior representation learning.

\subsubsection{Ablation Study. }
Training settings: To assess the effectiveness of the proposed training strategy, we conduct ablation studies on the MM-IMDB dataset under the following three settings:
\begin{itemize}
    \item \textbf{-D (Dual-Dimensional Attention Mechanism)}: Removes the dual-dimensional attent
    ion mechanism.
    \item \textbf{-G (Entity-Centroid Contrastive Learning)}: Removes the contrastive learning module; the fused features are not enhanced via contrastive learning.
    \item \textbf{-F (Adaptive Fusion)}: Replaces the adaptive fusion module with a static fusion strategy, where modality features are fused using a fixed rule.
\end{itemize}

Table 5 shows that replacing the adaptive fusion module with a static one causes a performance drop of over 5\% across all metrics, underscoring the importance of dynamic fusion. Ablation experiments also validated CL-DMDF's effective adaptation under noisy or missing modes.

Removing the dual-dimensional attention or contrastive learning module led to consistent performance drops, confirming their effectiveness in weighting and feature discrimination.

\begin{table}[t]
\centering
\begin{tabular}{lcc}
\toprule
\textbf{Model} & \textbf{MicroF1 (\%)} & \textbf{MacroF1 (\%)} \\
\midrule
CL-DMDF(ours) & 63.25 & 53.28 \\
CL-DMDF-D & 63.21 & 53.25 \\
CL-DMDF-D-G & 62.56 & 52.45 \\
CL-DMDF-D-G-F & 59.87 & 50.41 \\
\bottomrule
\end{tabular}
\caption{Ablation study results on the MM-IMDB dataset.}
\label{tab:ablation-mmimdb}
\end{table}

\subsubsection{Contrastive Learning Method Experiments. }
To evaluate the contrastive learning module, we conducted ablation studies by replacing it with existing methods, including MulCon~\cite{chih2022}, MCL~\cite{xu2022}, MLTC~\cite{wang2022}, and C-GMVAE~\cite{bai2022}. As shown in Table 6, our module consistently outperforms all baselines. Centroid-based type embeddings also outperform random initialization, capturing inter-class semantics and improving feature discrimination.

\begin{table}[t]
\centering
\begin{tabular}{lcc}
\toprule
\textbf{Model} & \textbf{MicroF1 (\%)} & \textbf{MacroF1 (\%)} \\
\midrule
MulCon & 62.14 & 52.03 \\
MCL & 62.38 & 52.26 \\
MLTC & 62.76 & 52.45 \\
C-GMVAE & 63.03 & 52.87 \\
\textbf{CL-DMDF(ours)} & \textbf{63.25} & \textbf{53.28} \\
\bottomrule
\end{tabular}
\caption{Comparison of CL-DMDF with contrastive learning baselines on the MM-IMDB dataset.}
\label{tab:contrastive-comparison}
\end{table}

\begin{figure}[t]
\centering
\includegraphics[width=1\columnwidth, height=5.5cm]{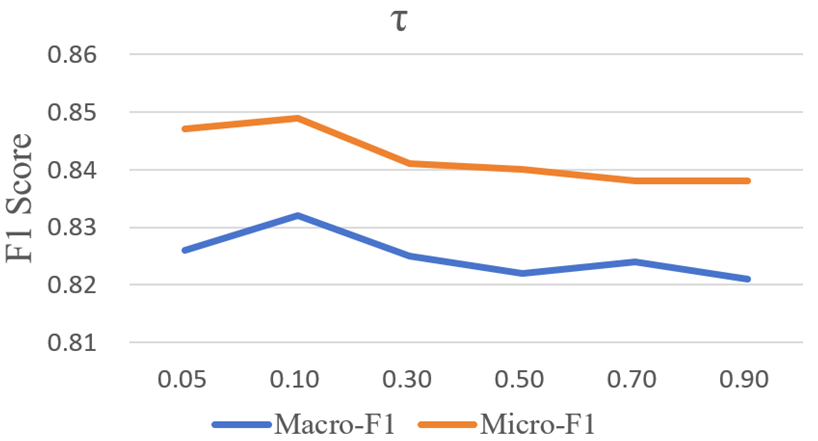} 
\caption{Effect of parameter $\tau$ on MM-IMDB.}
\label{fig5}
\end{figure}

\subsubsection{Parameter Analysis. }
In the entity-centroid contrastive module, the temperature parameter $\tau$ controls similarity scaling between positive and negative pairs, affecting embedding separation. A small $\tau$ overemphasizes hard negatives and distorts embeddings. Figure 5 shows that $\tau = 0.1$ achieves optimal results on the MM-IMDB dataset, with 83.2\% Macro-F1 and 84.9\% Micro-F1, enabling effective multimodal integration and improving downstream performance.

\subsubsection{Analysis of Dual-Dimensional Attention Mechanism. }
To assess the effectiveness of the dual-dimensional attention mechanism, we conduct a case study (Table 7) using three test samples \( A^r_1, A^r_2, A^r_3 \), in which semantically salient modalities are marked in red. The attention module assigns higher scores to informative modalities, validating its ability to identify modality relevance.

\begin{table}[t]
\centering
\small
\begin{tabular}{
    @{} >{\centering\arraybackslash}m{1.3cm}  % Poster 列
        >{\centering\arraybackslash}m{0.7cm}  % A^I
        >{\centering\arraybackslash}m{1.9cm}  % Plot 文本
        >{\centering\arraybackslash}m{0.7cm}  % A^T
        >{\centering\arraybackslash}m{0.8cm}  % Number
        >{\centering\arraybackslash}m{0.9cm}  % A^K
    @{}}
\toprule
\textbf{Poster} & \textbf{$A^{I}$} & \textbf{Plot} & \textbf{$A^{T}$} & \textbf{Number} & \textbf{$A^{K}$}\\
\midrule
\includegraphics[width=1.2cm,height=1.8cm]{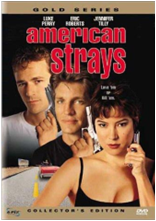} &
\textbf{0.824} &
The desert can be a lonely place for the \dots &
0.370 &
{\color{red}29} &
\textbf{0.822} \\
\midrule
\includegraphics[width=1.2cm,height=1.8cm]{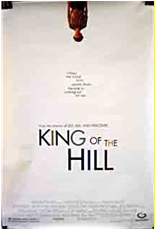} &
\textbf{0.373} &
{\color{red}A young boy struggles on his own in a run \dots} &
\textbf{0.708} &
{\color{red}38} &
\textbf{0.816} \\
\midrule
\includegraphics[width=1.2cm,height=1.8cm]{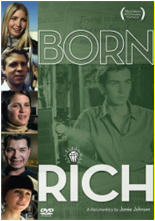} &
\textbf{0.659} &
A documentary directed by one of their own \dots &
\textbf{0.793} &
5 &
0.307 \\
\bottomrule
\end{tabular}
% \caption{Case study on dual-dimensional attention module.}
\caption{
Case study on dual-dimensional attention module.~\cite{li2023incorporating}. 
}
\label{tab:case_study}
\end{table}

As shown in Table 7, key modalities received high reliability scores, while less informative cues (e.g., emotion) scored lower, confirming the dual-dimensional attention effectively highlights salient modalities.For a fair comparison, we use the same test samples as those reported in IDKG~\cite{li2023incorporating}.

\section{Conclusion}
We propose Dynamic Multimodal Data Fusion model based on Contrastive Learning(CL-DMDF) that addresses semantic inconsistency and task-adaptive fusion. CL-DMDF integrates dual-dimensional attention for reliable weighting, centroid-based contrastive learning for discriminative representations, and an adaptive fusion module to balance accuracy and efficiency. Experiments on three benchmarks demonstrate consistent improvements over state-of-the-art methods. Future work will explore extensions to incomplete modalities and real-time scenarios.

\section{Acknowledgments}
This work was supported by the National Natural Science Foundation of China (62472204, 52574191, 62072220), the youth talent support program of `Xing Liao Talent Program' (XLYC2203003), the Basic Scientific Research Project of the Department of Education of Liaoning Province (LJ232510140001).

\bigskip

\bibliography{aaai2026}

\begin{thebibliography}{39}
\providecommand{\natexlab}[1]{#1}

\bibitem[{Amir et~al.(2018)Amir, Liang, Poria et~al.}]{amir2018}
Amir, Z.; Liang, P.~P.; Poria, S.; et~al. 2018.
\newblock Multi-Attention Recurrent Network for Human Communication Comprehension.
\newblock In \emph{Proceedings of the AAAI Conference on Artificial Intelligence}, 5642.

\bibitem[{Bai, Kong, and Gomes(2022)}]{bai2022}
Bai, J.~W.; Kong, S.~F.; and Gomes, C.~P. 2022.
\newblock Gaussian Mixture Variational Autoencoder with Contrastive Learning for Multi-Label Classification.
\newblock In \emph{Proceedings of the International Conference on Machine Learning}, 1383--1398.

\bibitem[{Bai et~al.(2019)Bai, Liang, Salakhutdinov et~al.}]{bai2019}
Bai, S.; Liang, P.~P.; Salakhutdinov, R.; et~al. 2019.
\newblock Multimodal Transformer for Unaligned Multimodal Language Sequences.
\newblock \emph{Proceedings of the Conference of the Association for Computational Linguistics}, 6558.

\bibitem[{Braz et~al.(2021)Braz, Teixeira, Pedrini et~al.}]{braz2021}
Braz, L.; Teixeira, V.; Pedrini, H.; et~al. 2021.
\newblock Image-text Integration Using a Multimodal Fusion Network Module for Movie Genre Classification.
\newblock In \emph{Proceedings of the 11th International Conference of Pattern Recognition Systems}, 200--205.

\bibitem[{Camille, Clément, and Laurent(2013)}]{camille2013}
Camille, C.; Clément, F.; and Laurent, N. 2013.
\newblock Indoor Semantic Segmentation Using Depth Information.
\newblock \emph{arXiv}.
\newblock ArXiv:1301.3572.

\bibitem[{Chen, Wang, and Zhang(2024)}]{chen2024pcam}
Chen, Y.; Wang, H.; and Zhang, J. 2024.
\newblock Progressive Cross-Modal Attention Mechanism for Hierarchical Multimodal Fusion.
\newblock In \emph{Proceedings of the IEEE/CVF Conference on Computer Vision and Pattern Recognition (CVPR)}, 11245--11254.

\bibitem[{Chih et~al.(2022)Chih, Pi, Ting et~al.}]{chih2022}
Chih, C.~H.; Pi, J.~T.; Ting, C.~Y.; et~al. 2022.
\newblock A Comprehensive Study of Spatiotemporal Feature Learning for Social Media Popularity Prediction.
\newblock In \emph{Proceedings of the 30th ACM International Conference on Multimedia}, 7130--7134.

\bibitem[{Dhruv et~al.(2019)Dhruv, Jaipal, Manish et~al.}]{dhruv2019a}
Dhruv, K.; Jaipal, S.~G.; Manish, G.; et~al. 2019.
\newblock MVAE: Multimodal Variational Autoencoder for Fake News Detection.
\newblock In \emph{Proceedings of the World Wide Web Conference}, 2915--2921.

\bibitem[{Ding, Sun, and Zhao(2023)}]{ding2023}
Ding, C.~Y.; Sun, S.~L.; and Zhao, J. 2023.
\newblock MST-GAT: A Multimodal Spatial–Temporal Graph Attention Network for Time Series Anomaly Detection.
\newblock \emph{Information Fusion}, 89: 527--536.

\bibitem[{Gao et~al.(2019)Gao, Jiang, You et~al.}]{gao2019}
Gao, P.; Jiang, Z.~K.; You, H.~X.; et~al. 2019.
\newblock Dynamic Fusion with Intra-and Inter-Modality Attention Flow for Visual Question Answering.
\newblock In \emph{Proceedings of the IEEE/CVF Conference on Computer Vision and Pattern Recognition}, 6639--6648.

\bibitem[{Hu et~al.(2021)Hu, Liu, Zhao et~al.}]{hu2021}
Hu, J.~W.; Liu, Y.~C.; Zhao, J.~M.; et~al. 2021.
\newblock MMGCN: Multimodal Fusion via Deep Graph Convolution Network for Emotion Recognition in Conversation.
\newblock In \emph{Proceedings of the 59th Annual Meeting of the Association for Computational Linguistics and the 11th International Joint Conference on Natural Language Processing}, 5666--5675.

\bibitem[{Hu et~al.(2020{\natexlab{a}})Hu, Feng, Sun et~al.}]{hu2020}
Hu, Z.~W.; Feng, G.; Sun, J.~Y.; et~al. 2020{\natexlab{a}}.
\newblock Bi-Directional Relationship Inferring Network for Referring Image Segmentation.
\newblock In \emph{Proceedings of the IEEE/CVF Conference on Computer Vision and Pattern Recognition}, 4424--4433.

\bibitem[{Hu et~al.(2020{\natexlab{b}})Hu, Feng, Sun et~al.}]{hu2020b}
Hu, Z.~W.; Feng, G.; Sun, J.~Y.; et~al. 2020{\natexlab{b}}.
\newblock Bi-Directional Relationship Inferring Network for Referring Image Segmentation.
\newblock In \emph{Proceedings of the IEEE/CVF Conference on Computer Vision and Pattern Recognition}, 4424--4433.

\bibitem[{Janani et~al.(2021)Janani, Li, Hamid et~al.}]{janani2021}
Janani, V.; Li, T.; Hamid, R.~H.; et~al. 2021.
\newblock Multimodal Deep Learning Models for Early Detection of Alzheimer’s Disease Stage.
\newblock \emph{Scientific Reports}, 11: 1--13.

\bibitem[{Jiang, Zhang, and Tan(2024)}]{jiang2024unifm}
Jiang, Y.; Zhang, R.; and Tan, J. 2024.
\newblock UniFM: Unified Multimodal Fusion with Cross-modal Token Sharing.
\newblock In \emph{Proceedings of the IEEE/CVF Conference on Computer Vision and Pattern Recognition (CVPR)}.

\bibitem[{Li et~al.(2023)Li, Qi, Zhang, Chen, Tan, Xia, and Tian}]{li2023incorporating}
Li, J.; Qi, G.; Zhang, C.; Chen, Y.; Tan, Y.; Xia, C.; and Tian, Y. 2023.
\newblock Incorporating Domain Knowledge Graph into Multimodal Movie Genre Classification with Self-Supervised Attention and Contrastive Learning.
\newblock In \emph{Proceedings of the 31st ACM International Conference on Multimedia}, 6953--6961. New York, NY, USA: Association for Computing Machinery.

\bibitem[{Li et~al.(2018)Li, Li, Kuo, Shen, and Jia}]{li2018}
Li, R.~Y.; Li, K.~C.; Kuo, Y.~C.; Shen, X.~Y.; and Jia, J. 2018.
\newblock Referring Image Segmentation via Recurrent Refinement Networks.
\newblock In \emph{Proceedings of the IEEE Conference on Computer Vision and Pattern Recognition}, 5745--5753.

\bibitem[{Li, Xu, and Liu(2025)}]{li2025tokenmixer}
Li, Z.; Xu, C.; and Liu, H. 2025.
\newblock MM-TokenMixer: Token Mixing for Efficient Multimodal Fusion.
\newblock In \emph{Proceedings of the AAAI Conference on Artificial Intelligence}.

\bibitem[{Liu, Zhao, and Zhang(2025)}]{liu2025exploring}
Liu, Q.; Zhao, Y.; and Zhang, W. 2025.
\newblock Exploring Adaptive Sparse Attention for Multimodal Fusion under Uncertainty.
\newblock In \emph{Proceedings of the AAAI Conference on Artificial Intelligence}, volume~39, 6993--7001.

\bibitem[{Liu, Fan, and Li(2025)}]{liu2025lvmf}
Liu, S.; Fan, Z.; and Li, M. 2025.
\newblock LVMF: Latent-Variable Multimodal Fusion with Uncertainty-Aware Representation Learning.
\newblock In \emph{Proceedings of the AAAI Conference on Artificial Intelligence}.
\newblock To appear.

\bibitem[{Liu et~al.(2019)Liu, Shi, Duan et~al.}]{liu2019}
Liu, Z.~Y.; Shi, S.; Duan, Q.~T.; et~al. 2019.
\newblock Salient Object Detection for RGB-D Image by Single Stream Recurrent Convolution Neural Network.
\newblock \emph{Neurocomputing}, 363: 46--57.

\bibitem[{Mateusz et~al.(2018)Mateusz, Carl, Adam et~al.}]{mateusz2018}
Mateusz, M.; Carl, D.; Adam, S.; et~al. 2018.
\newblock Learning Visual Question Answering by Bootstrapping Hard Attention.
\newblock In \emph{Proceedings of the European Conference on Computer Vision}, 3--20.

\bibitem[{Muralidhar et~al.(2019)Muralidhar, Jayakumar, Menick et~al.}]{siddhant2019}
Muralidhar, S.; Jayakumar; Menick, J.; et~al. 2019.
\newblock Multiplicative Interactions and Where to Find Them.
\newblock \url{https://openreview.net/pdf?id=rylnK6VtDH}.
\newblock Accessed: 2025-05-10.

\bibitem[{Nekrasov, Shuai-Hua, and Reid(2018)}]{vladimir2018}
Nekrasov, V.; Shuai-Hua, S.; and Reid, I. 2018.
\newblock Lightweight RefineNet for Real-Time Semantic Segmentation.
\newblock In \emph{Proceedings of the British Machine Vision Conference}, 125.

\bibitem[{Nihar, Kevin, and Peyman(2021)}]{nihar2021}
Nihar, B.; Kevin, D.; and Peyman, N. 2021.
\newblock Generalized Zero-Shot Learning Using Multimodal Variational Auto-Encoder with Semantic Concepts.
\newblock In \emph{Proceedings of the 2021 IEEE International Conference on Image Processing}, 1284--1288.

\bibitem[{Schlegel et~al.(2021)Schlegel, Lengerich, Weninger et~al.}]{daniel2021}
Schlegel, D.; Lengerich, B.; Weninger, T.; et~al. 2021.
\newblock Efficient RGB-D Semantic Segmentation for Indoor Scene Analysis.
\newblock In \emph{Proceedings of the International Conference on Robotics and Automation}, 13525--13531.

\bibitem[{Sudharsan, Diyi, and Soon(2021)}]{sankaran2021}
Sudharsan, S.; Diyi, Y.; and Soon, L. 2021.
\newblock Refining Multimodal Representations Using a Modality-Centric Self-Supervised Module.
\newblock \url{https://openreview.net/pdf?id=hB2HIO39r8G}.
\newblock Accessed: 2025-05-10.

\bibitem[{Sun et~al.(2021)Sun, Liu, Tao et~al.}]{sun2021}
Sun, L.~C.; Liu, B.; Tao, J.~H.; et~al. 2021.
\newblock Multimodal Cross-And Self-Attention Network for Speech Emotion Recognition.
\newblock In \emph{Proceedings of the ICASSP 2021-2021 IEEE International Conference on Acoustics, Speech and Signal Processing}, 4275--4279.

\bibitem[{Sun et~al.(2020)Sun, Sarma, Scherlis et~al.}]{sun2020}
Sun, Z.; Sarma, P.; Scherlis, W.; et~al. 2020.
\newblock Learning Relationships Between Text, Audio, And Video Via Deep Canonical Correlation for Multimodal Language Analysis.
\newblock In \emph{Proceedings of the AAAI Conference on Artificial Intelligence}, volume~34, 8992--8999.

\bibitem[{Wang, Dai et~al.(2022)}]{wang2022}
Wang, R.; Dai, X.; et~al. 2022.
\newblock Contrastive Learning-Enhanced Nearest Neighbor Mechanism for Multi-Label Text Classification.
\newblock In \emph{Proceedings of the 60th Annual Meeting of the Association for Computational Linguistics}, 672--679.

\bibitem[{Wang et~al.(2020)Wang, Huang, Sun et~al.}]{wang2020}
Wang, Y.; Huang, W.; Sun, F.; et~al. 2020.
\newblock Deep Multimodal Fusion by Channel Exchanging.
\newblock \emph{Advances in Neural Information Processing Systems}, 33: 4835--4845.

\bibitem[{Wang, Lin, and Song(2025)}]{wang2025modality}
Wang, Y.; Lin, J.; and Song, J. 2025.
\newblock Modality-Aware Reinforced Graph Fusion for Social Event Detection.
\newblock In \emph{Proceedings of the ACM International Conference on Multimedia (ACM MM)}, 234--245.

\bibitem[{Wu, Sun, and Huang(2024)}]{wu2024gnnadapter}
Wu, K.; Sun, X.; and Huang, Y. 2024.
\newblock GNN-Adapter: Plug-and-Play Graph Modules for Pretrained Multimodal Transformers.
\newblock In \emph{Proceedings of the 2024 Annual Meeting of the Association for Computational Linguistics (ACL)}.

\bibitem[{Xu, Feng, and Huang(2022)}]{xu2022}
Xu, K.~L.; Feng, M.; and Huang, W.~Q. 2022.
\newblock Seeing Speech: Magnetic Resonance Imaging-Based Vocal Tract Deformation Visualization Using Cross-Modal Transformer.
\newblock In \emph{Proceedings of the 30th ACM International Conference on Multimedia}, 6947--6949.

\bibitem[{Xue and Marculescu(2023)}]{xue2023dynamic}
Xue, Z.; and Marculescu, R. 2023.
\newblock Dynamic Multimodal Fusion.
\newblock In \emph{Proceedings of the IEEE/CVF Conference on Computer Vision and Pattern Recognition Workshops}, 2575--2584.

\bibitem[{Yang, Xu, and Gao(2020)}]{yang2020}
Yang, K.-C.; Xu, H.; and Gao, K. 2020.
\newblock CM-BERT: Cross-modal BERT for Text-Audio Sentiment Analysis.
\newblock In \emph{Proceedings of the 28th ACM International Conference on Multimedia}, 521--528.

\bibitem[{Yang, Tan, and Gao(2024)}]{yang2024dynamic}
Yang, M.; Tan, Z.; and Gao, L. 2024.
\newblock Dynamic Sparse Transformer for Robust Video-Text Fusion.
\newblock In \emph{Proceedings of the IEEE/CVF Conference on Computer Vision and Pattern Recognition (CVPR)}, 11720--11730.

\bibitem[{Yang et~al.(2023)Yang, Peng, Wang et~al.}]{yang2023}
Yang, X.; Peng, J.~W.; Wang, Z.~H.; et~al. 2023.
\newblock Transforming Visual Scene Graphs to Image Captions.
\newblock In \emph{Proceedings of the 61st Annual Meeting of the Association for Computational Linguistics}, 12427--12440.

\bibitem[{Zhang, Hu, and Tan(2025)}]{zhang2025lightfusion}
Zhang, W.; Hu, Z.; and Tan, F. 2025.
\newblock LightFusion: Efficient Transformer-based Multimodal Fusion for Real-Time Applications.
\newblock In \emph{Proceedings of the International Joint Conference on Artificial Intelligence (IJCAI)}.
\newblock To appear.

\end{thebibliography}

\end{document}